\pdfoutput=1
\documentclass[11pt]{article}
\usepackage{colortbl}
\usepackage[table]{xcolor}

\usepackage[preprint]{acl}
\usepackage{times}
\usepackage{latexsym}

\usepackage[T1]{fontenc}
\usepackage[utf8]{inputenc}

\usepackage{microtype}
\usepackage{amsmath,amssymb}
\usepackage{subcaption} % For subfigure environment

\usepackage{inconsolata}

\usepackage{setspace}

\usepackage{algorithm}
\usepackage{algorithmic}

\usepackage{graphicx}
\usepackage{times}
\usepackage{latexsym}
\usepackage[T1]{fontenc}
\usepackage[utf8]{inputenc}
\usepackage{pifont}
\newcommand{\cmark}{\ding{51}}%
\newcommand{\xmark}{\ding{55}}%
\newcommand{\smallstrut}{\rule{0pt}{0.7em}}
\usepackage{fancyvrb}
\usepackage{soul}
\usepackage{microtype}
\usepackage{tabularx}
\usepackage{xspace}
\usepackage{makecell}
\usepackage{tipa}
\usepackage{graphicx}
\usepackage{booktabs}
\usepackage{multirow}
\usepackage{caption}
\usepackage{amsmath} 
\usepackage{arydshln}
\usepackage{amssymb}  % For the \blacksquare symbol
\usepackage{xcolor}   % For the \color command
\usepackage{tcolorbox}
\usepackage{tabularx}
\usepackage{color}
\usepackage{bbm}
\usepackage{svg}
\usepackage{listings}
\usepackage{array} 
\usepackage{amsfonts}
\usepackage{marvosym}
\title{SemPA: Improving Sentence Embeddings of Large Language Models through Semantic Preference Alignment}

\author{
    Ziyang Chen\textsuperscript{*,1}, 
    Zhenxuan Huang\textsuperscript{*,1},
    Yile Wang\textsuperscript{\Letter,1}, 
    Weiqin Wang\textsuperscript{1}, 
    Lu Yin\textsuperscript{2}, 
    Hui Huang\textsuperscript{1} \\
    \textsuperscript{1}College of Computer Science and Software Engineering, Shenzhen University \\
    \textsuperscript{2}School of Computer Science and Electronic Engineering, University of Surrey
}

\begin{document}
\maketitle

% --- custom author footnotes ---
\renewcommand{\thefootnote}{*}
\footnotetext{Equal contribution.}
%\renewcommand{\thefootnote}{$\dagger$}
%\footnotetext{Work done during the internship at Shenzhen University.}
\renewcommand{\thefootnote}{\protect\Letter}
%\footnotetext{Corresponding to \texttt{wangyile@szu.edu.cn}}
\footnotetext{Corresponding author: \{\texttt{wangyile@szu.edu.cn}\}.}
% restore normal numeric footnotes for the paper body
\setcounter{footnote}{0}
\renewcommand{\thefootnote}{\arabic{footnote}}

\begin{abstract}
Traditional sentence embedding methods employ token-level contrastive learning on non-generative pre-trained models. Recently, there have emerged embedding methods based on generative large language models (LLMs). These methods either rely on fixed prompt templates~\cite{jiang-etal-2024-scaling} or involve modifications to the model architecture~\cite{behnamghader2024llmvec}. The former lacks further optimization of the model and results in limited performance, while the latter alters the internal computational mechanisms of the model, thereby compromising its generative capabilities. We propose \textsc{SemPA}, a novel approach that boosts the sentence representations while preserving the generative ability of LLMs via semantic preference alignment. We leverage sentence-level Direct Preference Optimization (DPO) to efficiently optimize LLMs on a paraphrase generation task, where the model learns to discriminate semantically equivalent sentences while preserving inherent generative capacity. Theoretically, we establish a formal connection between DPO and contrastive learning under the Plackett-Luce model framework. Empirically, experimental results on both semantic textual similarity tasks and various benchmarks for LLMs show that \textsc{SemPA} achieves better semantic representations without sacrificing the inherent generation capability of LLMs. We release the code at \texttt{\url{https://github.com/szu-tera/SemPA}}.

\end{abstract}

\section{Introduction}

\begin{figure}[t!]
	\centering
	\includegraphics[width=0.98\linewidth]{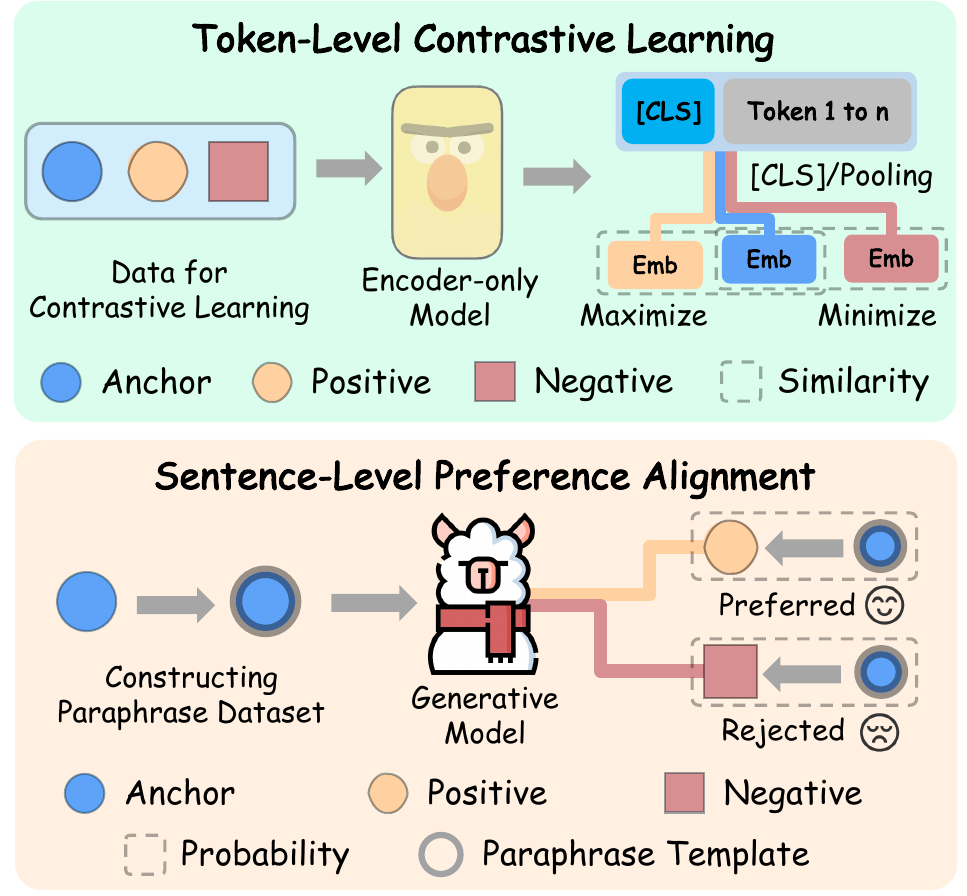}
	\caption{Comparison of sentence embedding methods. Top: contrastive learning for encoder-only models. Bottom: our semantic preference alignment for LLMs.}
	\label{fig:intro}
\end{figure}

Representing text as a semantically meaningful vector is a fundamental and essential step in natural language processing, supporting various downstream applications such as text classification, clustering, and information retrieval. Seminal works primarily leverage token-level contrastive learning~\cite{gao-etal-2021-simcse,chuang-etal-2022-diffcse} and derive sentence embeddings by extracting the hidden states of special tokens such as \texttt{[CLS]} or the pooled representation of all tokens. The success of such methods relies on powerful pre-trained encoder-only models like BERT~\cite{devlin-etal-2019-bert} or RoBERTa~\cite{liu2019roberta}, as shown in Figure~\ref{fig:intro}(top).

\begin{table*}[t!]
\centering
\resizebox{1\linewidth}{!}{
\begin{tabular}{lcccc}
\toprule
\textbf{Related Works}  & \textbf{Models}& \textbf{Methods} & \textbf{Optimization Levels}& \textbf{Generation}  \\ 
\midrule
SimCSE~\cite{gao-etal-2021-simcse}& PTMs& Contrastive Learning& Token-Level& \textcolor{red}{\xmark} \\
LLM2Vec~\cite{behnamghader2024llmvec}& LLMs & Contrastive Learning& Token-Level & \textcolor{red}{\xmark}\\ 
NV-Embed~\cite{lee2025nvembed} & LEMs & Contrastive Learning & Token-Level & \textcolor{red}{\xmark}\\ 
Qwen3-embedding~\cite{zhang2025qwen3embeddingadvancingtext} & LEMs & Contrastive Learning & Token-Level & \textcolor{red}{\xmark}\\ 
\midrule
PromptBERT~\cite{jiang-etal-2022-promptbert}& PTMs& Prompt Learning& Context-Level& \textcolor{red}{\xmark} \\
PromptEOL~\cite{jiang-etal-2024-scaling} & LLMs& Prompt Learning& Context-Level& \textcolor{teal}{\cmark} \\ 
MetaEOL~\cite{lei-etal-2024-meta}& LLMs& Prompt Learning& Context-Level& \textcolor{teal}{\cmark} \\ 
Contrastive Prompting~\cite{cheng-etal-2025-contrastive} & LLMs & Prompt Learning& Context-Level & \textcolor{red}{\xmark}\\ 
\midrule
Token Prepending~\cite{fu-etal-2025-token}& LLMs & Internal Operations Intervention& Architecture-Level & \textcolor{red}{\xmark}\\ 
Hierarchical Token Prepending~\cite{ding2025hierarchicaltokenprependingenhancing}& LLMs & Internal Operations Intervention& Architecture-Level & \textcolor{red}{\xmark}\\
\midrule
INSTRUCTOR~\cite{su-etal-2023-one}& PTMs& Instruction Tuning& Sentence-Level& \textcolor{red}{\xmark}\\ 
GRIT~\cite{muennighoff2025generative}& LLMs& Instruction Tuning& Sentence-Level& \textcolor{teal}{\cmark}\\ 
\textbf{\textsc{SemPA} (Ours)}  & \textbf{LLMs} & \textbf{Preference Alignment}& \textbf{Sentence-Level} & \textcolor{teal}{\cmark}\\ 
\bottomrule
\end{tabular}}
\caption{Comparison of typical sentence embedding related works and our \textsc{SemPA}. PTMs, LLMs, and LEMs denote pre-trained models, large language models, and large embedding models, respectively.}
\label{tab:relatedwork}
\end{table*}

Generative large language models (LLMs; \citealp{brown2020language,openaichatgpt,achiam2023gpt}), with massive parameters and trained on extensive data, have demonstrated remarkable capabilities. This has led to the emergence of sentence embedding methods based on either prompting~\cite{jiang-etal-2022-promptbert,lei-etal-2024-meta} or modifying these LLMs~\cite{behnamghader2024llmvec,fu-etal-2025-token}. However, prompting-based approaches, which rely solely on prompt engineering, have inherent performance limitations. While model-modification methods can improve sentence representations, they often compromise LLMs’ generative ability.

To achieve semantic representation with LLMs while preserving their inherent generative capabilities, we fine-tune LLMs using preference alignment~\cite{NEURIPS2022_b1efde53}. Preference alignment was originally designed to enhance the truthfulness and safety of LLMs by constructing preference data pairs, enabling sentence-level optimization in certain aspects. Inspired by this, we propose \textsc{SemPA}, a semantic preference alignment method that operates at the sentence level by applying Direct Preference Optimization (DPO; \citealp{NEURIPS2023_a85b405e}) to a paraphrase generation task. This allows LLMs to learn fine-grained semantic distinctions by assigning higher relative probabilities to semantically accurate paraphrases and thus improves the underlying semantic representations with lightweight training, without sacrificing their generative proficiency, as shown in Figure~\ref{fig:intro}(bottom).

Theoretically, we provide a derivation that connects contrastive learning and DPO in a unified form, offering a new perspective on why preference alignment serves as an effective paradigm for representation learning. Empirically, results on Semantic Textual Similarity (STS) tasks and various generative benchmarks (\citealp{cobbe2021training},~\citealp{hendrycks2021measuring}, etc.) using LLaMA models~\cite{touvron2023llama2openfoundation,grattafiori2024llama3herdmodels} demonstrate that \textsc{SemPA} outperforms pre-trained and prompt-based baselines and stays competitive with specialized embedding models, while maintaining the core generation capability of LLMs across various tasks.

\section{Related Work}

\noindent\textbf{Sentence Embedding.}
Sentence embedding aims to encode the semantic content of text into numerical vectors for semantic similarity assessment and downstream tasks. Current sentence embedding methods can be categorized into four types:

1) \textit{Token-Level.} Token-level contrastive learning is widely used in traditional pre-trained models~\cite{gao-etal-2021-simcse,chuang-etal-2022-diffcse} or generative language models~\cite{neelakantan2022text,behnamghader2024llmvec} for sentence embedding. It optimizes the token-level representations of models by distinguishing positive and negative samples, thereby extracting general semantic representations from special tokens such as \texttt{[CLS]} or the pooling result of all tokens. Contrastive learning is also extensively employed in commercial embedding models such as NV-Embed~\cite{lee2025nvembed}, Qwen3-embedding~\cite{zhang2025qwen3embeddingadvancingtext}, and Gemini embedding~\cite{lee2025gemini}, demonstrating strong performance and scalability.

2) \textit{Sentence-Level.} To further adapt LLMs to various downstream tasks (e.g., classification, retrieval, and clustering), \citet{su-etal-2023-one} and \citet{muennighoff2025generative} employ sentence-level instruction tuning to train LLMs to acquire task-specific semantic representations. These types of methods often rely on large-scale and high-quality instruction datasets such as MEDI~\cite{su-etal-2023-one} and E5~\cite{wang2022text}.

3) \textit{Context-Level.} There have been some studies on context-based modification that extract overall semantic representations of given texts without fine-tuning the model parameters. \citet{jiang-etal-2022-promptbert} obtains semantic information based on the \texttt{[MASK]} token of BERT through specific templates. \citet{jiang-etal-2024-scaling} and \citet{lei-etal-2024-meta} design prompts to extract semantic information from generative models such as LLaMA. \citet{cheng-etal-2025-contrastive} enhances semantic representation by leveraging multiple contexts with different meanings. The effectiveness of such context-based methods largely depends on the model’s inherent capabilities and the quality of specialized prompts.

4) \textit{Architecture-Level.} Modifying the internal structure of LLMs has emerged recently for sentence embeddings~\cite{fu-etal-2025-token,ding2025hierarchicaltokenprependingenhancing}. These approaches alter the computational operations within the Transformer decoder architecture to extract specific hidden states as semantic representations. However, such invasive modifications change the model's structure, thus compromising the model’s original generative capabilities.

\noindent\textbf{Preference Alignment of LLMs.} Preference alignment has been shown to be an effective way to improve the safety and truthfulness of LLMs. It can be achieved by training LLMs on datasets of human or AI-generated preferences, using methods such as reinforcement learning from human feedback (RLHF; \citealp{NEURIPS2022_b1efde53}) or direct preference optimization (DPO; \citealp{NEURIPS2023_a85b405e}). Preference alignment optimizes the model beyond token-level prediction and improves the overall response quality, and it has also been applied to tasks such as combinatorial optimization~\cite{pan2025preference}, code generation~\cite{zhang-etal-2025-focused}, and mathematical reasoning~\cite{ICLR2025_31a57804}. To our knowledge, we are the first to explore preference alignment for improving the semantic representations of LLMs.

We summarize and compare the main related works with our \textsc{SemPA} in Table~\ref{tab:relatedwork}. Note that the primary goal of our work is not to compete with state-of-the-art embedding models in terms of performance, but rather to offer insights for improving semantic representations of LLMs through the perspective of lightweight preference alignment, which has not been explored in existing work.

\section{Preliminary}
\noindent\textbf{PromptEOL.} Given a sentence $\mathcal{S} = s_1, \ldots, s_n$, PromptEOL~\cite{jiang-etal-2024-scaling} extracts the sentence embedding of $\mathcal{S}$ using LLMs by inserting $\mathcal{S}$ in a template $\mathcal{T}$ defined as:
\[
\text{\textit{This sentence: ``$\mathcal{S}$'' means in one word:``}}
\]

The template with filled sentence $\mathcal{T}(\mathcal{S})$ is used as the input for a generative model $\mathcal{M}$, and the corresponding sentence embedding of $\mathcal{S}$ is the hidden state of the last token in the final layer $L$:

\begin{equation}
\begin{aligned}
{h^L_1}, {h^L_2}, ..., {h^L_{\rm last}} &= \mathcal{M}(\mathcal{T}(\mathcal{S})),\\
{\rm Emb}(\mathcal{S}) &= {h^L_{\rm last}},
\end{aligned}
\label{eq:prompteol}
\end{equation}
where ${\rm Emb}$($\mathcal{S}$) serves as a training-free baseline embedding extracted from generative LLMs.

\noindent\textbf{Direct Preference Optimization.}
To tailor the behavior of model $\mathcal{M}$ more closely to human preferences, DPO~\cite{NEURIPS2023_a85b405e} directly optimizes a policy by leveraging a closed-form mapping from reward functions to optimal policies, thereby avoiding the need for an explicit reward model.
%DPO公式
Formally, given dataset $\mathcal{D}$ consisting of triplets $(x,y_w,y_l)$ where $y_w$ is preferred over $y_l$ for input $x$, the loss function of DPO is defined as:

\begin{equation}
\scalebox{0.65}{$
        \mathcal{L}_{\text{DPO}}(\pi_\theta; \pi_{\text{ref}}) = -\mathbb{E}_{(x, y_w, y_l) \sim \mathcal{D}}\Bigg[ \log \sigma \Big( \beta \log \frac{\pi_\theta(y_w \mid x)}{\pi_{\text{ref}}(y_w \mid x)} - \beta \log \frac{\pi_\theta(y_l \mid x)}{\pi_{\text{ref}}(y_l \mid x)} \Big) \Bigg],
    $}
    \label{eq:dpo}
\end{equation}
where $\pi_\theta$ is the current policy to be optimized and $\pi_{\text{ref}}$ is the reference policy by the original model $\mathcal{M}$.
After identity transformation, the above formula can be expressed as:

\begin{equation}
\scalebox{0.7}{$
        \mathcal{L}_{\text{DPO}}(\pi_\theta; \pi_{\text{ref}}) =
         -\mathbb{E}_{(x, y_w, y_l) \sim \mathcal{D}}   
        \Bigg[ \log \frac{e^{\beta \log \frac{\pi_\theta(y_w \mid x)}{\pi_{\text{ref}}(y_w \mid x)}}}{e^{\beta \log \frac{\pi_\theta(y_w \mid x)}{\pi_{\text{ref}}(y_w \mid x)}}+e^{\beta \log \frac{\pi_\theta(y_l \mid x)}{\pi_{\text{ref}}(y_l \mid x)}}} \Bigg],
$}
\label{eq:dpo_transform}
\end{equation}
which shares a unified framework with contrastive learning, as discussed in Section~\ref{subsec:dpo_contrastive_link}.

\section{Our Method}

\begin{figure*}[t!]
	\centering
	\includegraphics[width=0.98\linewidth]{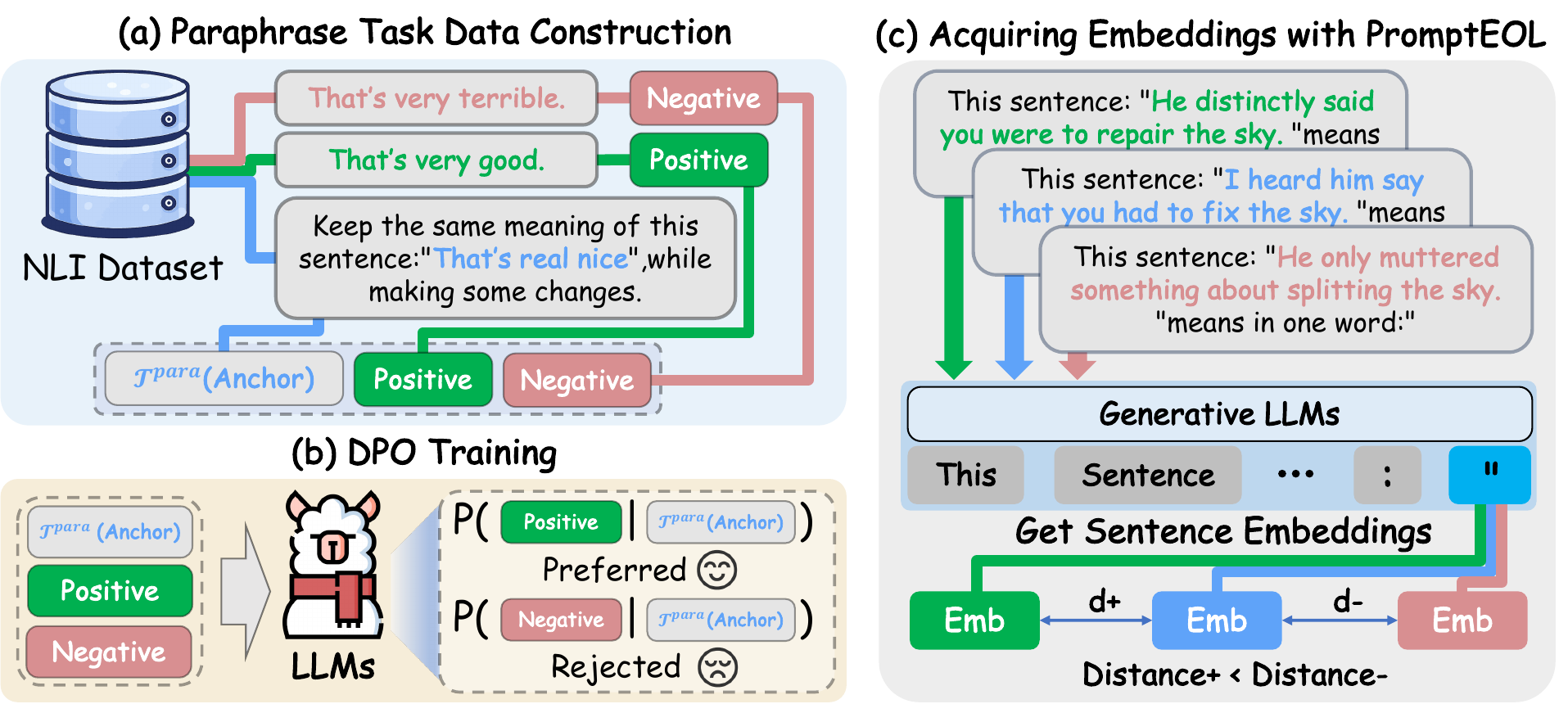}
	\caption{The overall pipeline of our proposed \textsc{SemPA} method. (a) We first construct the paraphrase generation preference pairs using NLI datasets (\S\ref{subsec:paraphrase}). (b) Then we perform semantic DPO training on LLMs (\S\ref{subsec:dpotraining}). (c) The final sentence embedding is acquired using the PromptEOL template (\S\ref{subsec:acquisition}).}
	\label{fig:SemPA}
\end{figure*}

\subsection{Paraphrase Generation Task}
\label{subsec:paraphrase}

We integrate the concept of ``paraphrase'' to improve the semantic representations of models. Traditional paraphrase-related tasks center on paraphrase identification~\cite{yin-schutze-2015-convolutional,wang-etal-2021-knowledge-guided,peng-etal-2022-predicate}, where an auto-encoding model is trained with an explicit classification head to determine whether two given sentences convey the same meaning. To leverage paraphrase to enhance the semantic representations of generative LLMs, we propose a paraphrase generation task and construct paraphrase-based preference pairs for model optimization.

Preference alignment pairs for training LLMs are typically sourced from user-annotated~\cite{NEURIPS2022_b1efde53} or model-generated~\cite{alpaca} data, which entail significant annotation costs or computational overhead. In this work, we employ existing natural language inference (NLI; \citealp{williams-etal-2018-broad}) datasets as the source of semantic preference pairs, thereby enabling a lightweight and low-cost model fine-tuning approach.

For instance, in NLI datasets, given a premise, another sentence needs to be labeled with one of the following labels: entailment, neutral, or contradiction. Following \citet{gao-etal-2021-simcse}, we treat the premise as the anchor sentence $x^*$, regard the entailment-labeled sentence as the positive sample $y^+$, and the contradiction-labeled sentence as the negative sample $y^-$, as shown in Figure~\ref{fig:SemPA}(a). The paraphrase generation task encourages the model to generate $y^+$ given $x^*$, which requires the generative model to possess fine-grained semantic understanding of the sentences.

\begin{table}[t!]  % 改用table（半页无需table*）
\centering
\setlength{\tabcolsep}{2pt}  % 进一步缩小列间距
\renewcommand{\arraystretch}{1.5}  % 适度降低行高，适配半页
\resizebox{0.48\textwidth}{!}{  % 核心：缩放到半页宽度（0.48避免溢出）
\begin{tabular}{lcc}
\toprule
& \textbf{Contrastive Learning} & \textbf{DPO} \\
% \frac{\exp\left(s_\theta(x, y_i)\right)}{\sum_{j=1}^{N} \exp\left(s_\theta(x, y_j)\right)}
\midrule
\textbf{Unified Form} & \multicolumn{2}{c}{$P(y_i {\text { is the best} }\mid x) = \frac{\exp\left(s_\theta(x, y_i)\right)}{\sum_{j=1}^{N} \exp\left(s_\theta(x, y_j)\right)}$}\\
\textbf{Candidate Set} & $y^+$ and $(y_1^-,\dots,y_{N-1}^-)$& $y_w$ and $y_l$  \\
\textbf{Ranking Event} & $y^+$ is the best & $y_w$ is the best \\
\textbf{Target Model} & auto-encoding model $e_{\theta}$& auto-regressive policy $\pi_{\theta}$, $\pi_{\text{ref}}$\\
\textbf{Scoring Function} $s_\theta$ & ${\rm sim}({e_{\theta}}(x),{e_{\theta}}(y))$  & $\beta \log \frac{{\pi_\theta}(y|x)}{{\pi_{{\text{ref}}}}(y|x)}$\\

\textbf{Objective Function} & $\frac{\exp(s_\theta(x,y^+))}{\exp(s_\theta(x,y^+))+\sum_{k=1}^{N-1}\exp(s_\theta(x,y_k^-))}$ & 
$\frac{\exp(s_\theta(x,y_w))}{\exp(s_\theta(x,y_w))+\exp(s_\theta(x,y_l))}$ \\
%\textbf{Optimization} & one-step update & step-by-step update\\
\bottomrule
\end{tabular}
}
\caption{Comparison between contrastive learning and DPO under the unified Plackett–Luce model framework.}
\label{tab:dpo_infonce_comp}  % 规范标签命名
\end{table}

\begin{table*}[t]
\centering
\scalebox{0.73}{
\begin{tabular}{lccccccccc}
\toprule
\textbf{Methods} & \textbf{Params} & \textbf{STS12} & \textbf{STS13} & \textbf{STS14} & \textbf{STS15} & \textbf{STS16} & \textbf{STS-B} & \textbf{SICK-R} & \textbf{Avg.} \\
\midrule
BERT-avg~\cite{devlin-etal-2019-bert} & 110M & 30.87 & 59.89 & 47.73 & 60.29 & 63.73 & 47.29 & 58.22 & 52.57 \\
Sentence-T5-avg~\cite{ni-etal-2022-sentence} & 4.8B & 34.97 & 60.19 & 47.59 & 66.40 & 70.62 & 62.83 & 63.57 & 58.02 \\
PromptBERT~\cite{jiang-etal-2022-promptbert} & 110M & 60.96 & 73.83 & 62.18 & 71.54 & 68.68 & 70.60 & 67.16 & 67.85 \\
SBERT-NLI-base~\cite{reimers-gurevych-2019-sentence} & 110M & 70.97 & 76.53 & 73.19 & 79.09 & 74.30 & 77.03 & 72.91 & 74.86 \\
SBERT-NLI-large~\cite{reimers-gurevych-2019-sentence} & 340M & 72.27 & 78.46 & 74.90 & 80.99 & 76.25 & 79.23 & 73.75 & 76.55 \\
LLM2Vec-LLaMA2~\cite{behnamghader2024llmvec} & 7B & 65.39 & 79.26 & 72.98 & 82.72 & 81.02 & 78.32 & 71.77 & 75.92 \\
% LLM2Vec-Mistral~\cite{behnamghader2024llmvec} & 7B & 67.65 & 83.90 & 76.97 & 83.80 & 81.91 & 80.42 & 75.55 & 78.60 \\
\hline
\specialrule{\lightrulewidth}{0pt}{0pt}
\rowcolor{gray!15}
\multicolumn{10}{c}{\textit{LLaMA2-7B}}\\
Mean Pooling & 7B & 35.45 & 53.62 & 39.87 & 54.68 & 53.00 & 41.52 & 48.86 & 46.71 \\
Echo Embedding~\cite{springer2025repetition} & 7B & 50.43 & 73.77 & 61.61 & 73.21 & 70.80 & 68.67 & 65.11 & 66.23 \\
PromptEOL~\cite{jiang-etal-2024-scaling} & 7B & 58.81 & 77.01 & 66.34 & 73.22 & 73.56 & 71.66 & 69.64 & 70.03 \\
Contrastive Prompting~\cite{cheng-etal-2025-contrastive} & 7B & 63.34 & 82.15 & 71.73 & 79.68 & 77.23 & 78.71 & 74.04 & 75.27 \\
Token Prepending~\cite{fu-etal-2025-token}  & 7B & 66.90 & 83.12 & 74.31 & 79.87 & \textbf{80.03} & \textbf{80.67} & 75.40 & 77.19 \\
\textbf{\textsc{SemPA} (Ours)} & 7B & \textbf{68.77} & \textbf{83.73} & \textbf{74.73} & \textbf{82.08} & 77.48 & 79.76 & \textbf{77.26} & \textbf{77.69} \\
\hline
\specialrule{\lightrulewidth}{0pt}{0pt}
\rowcolor{gray!15}
\multicolumn{10}{c}{\textit{LLaMA3-8B}}\\
Mean Pooling & 8B & 32.62 & 52.65 & 42.86 & 58.07 & 54.28 & 46.25 & 50.57 & 48.19 \\
Echo Embedding~\cite{springer2025repetition} & 8B & 50.30 & 71.75 & 62.43 & 72.01 & 75.62 & 69.75 & 65.79 & 66.81 \\
PromptEOL~\cite{jiang-etal-2024-scaling} & 8B & 60.89 & 78.57 & 68.18 & 76.75 & 77.16 & 72.83 & 68.94 & 71.90 \\
Contrastive Prompting~\cite{cheng-etal-2025-contrastive} & 8B & 60.80 & 80.67 & 70.06 & 78.34 & 77.46 & 75.05 & 70.28 & 73.24 \\
Token Prepending~\cite{fu-etal-2025-token} & 8B & 64.45 & 81.61 & 71.52 & 78.97 & \textbf{80.18} & 77.43 & 73.22 & 75.34 \\
\textbf{\textsc{SemPA} (Ours)} & 8B & \textbf{68.55} & \textbf{84.62} & \textbf{75.89} & \textbf{82.11} & 78.36 & \textbf{80.04} & \textbf{77.12} & \textbf{78.10} \\
\bottomrule
\end{tabular}}
\caption{Results on semantic textual similarity tasks, reported by Spearman correlation $\rho$$\times$$100$.}
\label{tab:sts-results}
\end{table*} 
\subsection{Semantic Preference Alignment via DPO}
\label{subsec:dpotraining}
Given the triplet ($x^*$, $y^+$, $y^-$), instead of directly performing sequence-to-sequence generation from input $x^*$ to output $y^+$, we align the model to prefer generating $y^+$ over $y^-$ conditioned on $x^*$, thereby preventing the model from degenerating into a paraphrasing model. We first insert the anchor $x^*$ into a paraphrasing instruction template $\mathcal{T}^{\rm para}$ as follows:
\begin{quote}
\textit{Keep the same meaning of this sentence: ``$x^*$'', while making some changes.}
\end{quote}
Then we apply the constructed data to DPO training, as shown in Figure~\ref{fig:SemPA}(b). In particular, we fill the template $\mathcal{T}^{\rm para}$ with the anchor sentence $x^*$ to obtain $\mathcal{T}^{\rm para}(x^*)$, which is used as the model input $x$ in Eq.~\ref{eq:dpo}; the positive sample $y^+$ serves as the preferred response $y_w$, and the negative sample $y^-$ serves as the rejected response $y_l$. We compare different paraphrasing templates $\mathcal{T}^{\rm para}$ in Section~\ref{subsec:template}.

We refer to the above alignment method as \textit{semantic preference alignment} (i.e., \textsc{SemPA}), which aims to: 1) retain the original architecture of the generative LLMs without modification; 2) enhance the semantic understanding capability of LLMs; and 3) preserve the foundational capabilities of the original LLMs through lightweight fine-tuning.

\subsection{Acquisition of Sentence Embedding}
\label{subsec:acquisition}
As shown in Figure~\ref{fig:SemPA}(c), after obtaining the semantic preference aligned model $\mathcal{M}^\textsc{\textsc{SemPA}}$, we follow PromptEOL by using the template $\mathcal{T}$ to extract our sentence embedding of $\mathcal{S}$ from the last-layer hidden state of the LLM similar to Eq.~\ref{eq:prompteol}:
\begin{equation}
\begin{aligned}
{h^L_1}, {h^L_2}, ..., {h^L_{\rm last}} &= \mathcal{M}^\textsc{\textsc{SemPA}}(\mathcal{T}(\mathcal{S})),\\
{\rm Emb}^{\textsc{SemPA}}(\mathcal{S}) &= {h^L_{\rm last}}
\end{aligned}
\label{eq:ours}
\end{equation}
We discuss different templates $\mathcal{T}$ to obtain the final sentence embedding ${\rm Emb}^{\textsc{SemPA}}(\mathcal{S})$ in Section~\ref{subsec:template}.

\subsection{Unifying Contrastive Learning and DPO}
\label{subsec:dpo_contrastive_link}
We offer a theoretical connection between contrastive learning and DPO in terms of their objective formulations, which can help us understand that DPO can also enhance semantic representations similar to contrastive learning. Specifically, they can both be formally regarded as specific instantiations of the Plackett-Luce model~\cite{af5079a1-8ca5-3727-a405-0a82390327b7}. For $N$ responses $\{y_1, \dots, y_N\}$, let the observed ranking be $\sigma = (y_{(1)} \succ y_{(2)} \succ \dots \succ y_{(N)})$, where $y_{(1)}$ is the optimal response and $y_{(N)}$ is the worst. The probability of the complete ranking is:
\begin{equation}
    P(\sigma \mid x) = \prod_{k=1}^{N} \frac{\exp\left(s_\theta(x, y_{(k)})\right)}{\sum_{j=k}^{N} \exp\left(s_\theta(x, y_{(j)})\right)},
\end{equation}
where $s_\theta$ is a score function between the input and the candidate. The above formula can be interpreted as a sequential selection process: the top-ranked candidate is chosen first, followed by the second-ranked candidate from the remaining options, and this procedure continues iteratively. 

Similarly, both contrastive learning and DPO focus solely on the selection probability of the top-ranked candidate. As a result, the formula can be simplified to:
\begin{equation}
    P(y_1 \text{ is the best} \mid x) = \frac{\exp\left(s_\theta(x, y_1)\right)}{\sum_{j=1}^{N} \exp\left(s_\theta(x, y_j)\right)}, 
\end{equation}
where the main difference lies in the scoring function $s_\theta$ and the final objective, as shown in Table~\ref{tab:dpo_infonce_comp}.

For the scoring function, contrastive learning uses a similarity function with representations (usually from the special token \texttt{[CLS]}) given by the encoder $e_{\theta}$, while DPO uses the log ratio of the policy $\pi_\theta$ to a generative reference model $\pi_{\text{ref}}$. For the objective $\mathcal{L}$, contrastive learning uses the log-likelihood InfoNCE~\cite{oord2018representation} similar to Eq.~\ref{eq:dpo_transform} in DPO.

\section{Experiments}
\subsection{Settings}
\noindent\textbf{Datasets.}
We use the natural language inference dataset for training, which is also used in SimCSE\footnote{\url{https://huggingface.co/datasets/princeton-nlp/datasets-for-simcse/blob/main/nli_for_simcse.csv}}. This dataset consists of 275K triplets, and we use a subset of 40$\sim$80K samples based on the performance on the development set.

For evaluation, we use seven standard Semantic Textual Similarity (STS) datasets, including STS 2012-2016~\cite{agirre-etal-2012-semeval,agirre-etal-2013-sem,agirre-etal-2014-semeval,agirre-etal-2015-semeval,agirre-etal-2016-semeval}
, STS-B~\cite{cer-etal-2017-semeval}, and SICK-R~\cite{marelli-etal-2014-sick}.

\noindent\textbf{Baselines.} 
Following \citet{fu-etal-2025-token}, we compare our \textsc{SemPA} with the following baselines: \textbf{BERT-avg} \cite{devlin-etal-2019-bert} averages the last-layer token embeddings from BERT to form a sentence embedding. \textbf{Sentence-T5-avg} \cite{ni-etal-2022-sentence} uses the mean pooling of token representations from the T5 encoder. \textbf{PromptBERT}~\cite{jiang-etal-2022-promptbert} represents sentences with the \texttt{[MASK]} token using pre-defined prompts. \textbf{SBERT}~\cite{reimers-gurevych-2019-sentence} fine-tunes a BERT model with siamese networks and refines its semantic representations. \textbf{LLM2Vec}~\cite{behnamghader2024llmvec} converts decoder-only LLMs into bidirectional encoders by modifying the attention pattern and performs contrastive learning. \textbf{Echo Embedding}~\cite{springer2025repetition} forms sentence representations by repeating the input twice and pooling only the hidden states from the second occurrence. \textbf{PromptEOL}~\cite{jiang-etal-2024-scaling} designs prompts and extracts the last-token hidden state as semantic representation from LLMs. \textbf{Contrastive Prompting} \cite{cheng-etal-2025-contrastive} introduces an auxiliary prompt to contrastively intervene in the original prompt’s representations, thereby suppressing redundant information. \textbf{Token Prepending} \cite{fu-etal-2025-token} prepends the per-layer sentence embedding to the next layer’s input, allowing early tokens to attend to full-sentence context under causal attention. For the Contrastive Prompting and Token Prepending baselines, PromptEOL is also applied for embedding extraction and fair comparison. 

\begin{figure}[t!]
    \centering
    \includegraphics[width=\linewidth]{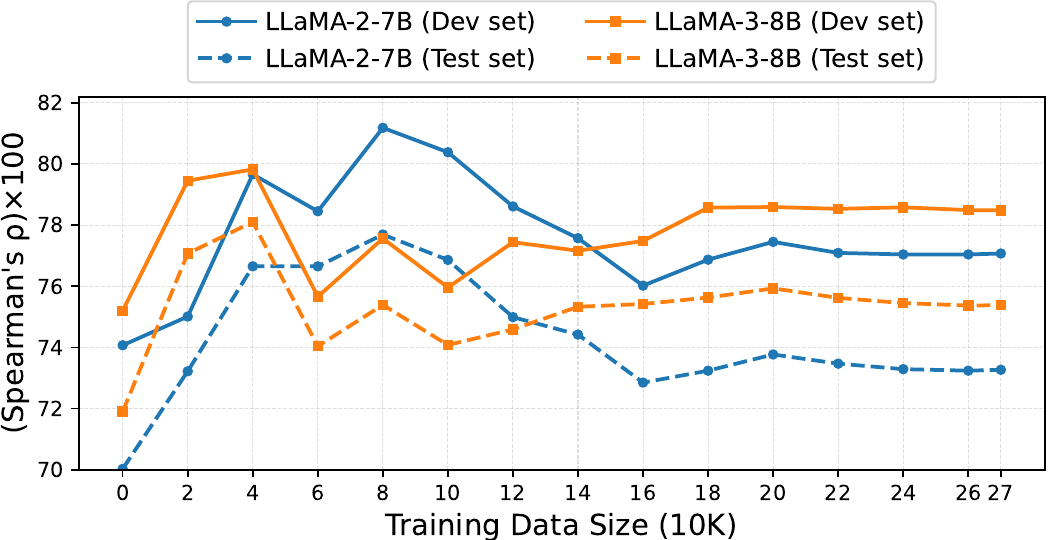}
    \caption{Performance of LLaMA models on the STS-B dev set and STS test sets with different training data sizes.
STS test results are averaged over all STS tasks.}
    \label{fig:SP}
\end{figure}
\begin{figure}[t!]
    \centering
    \resizebox{0.5\textwidth}{!}{
   \includegraphics{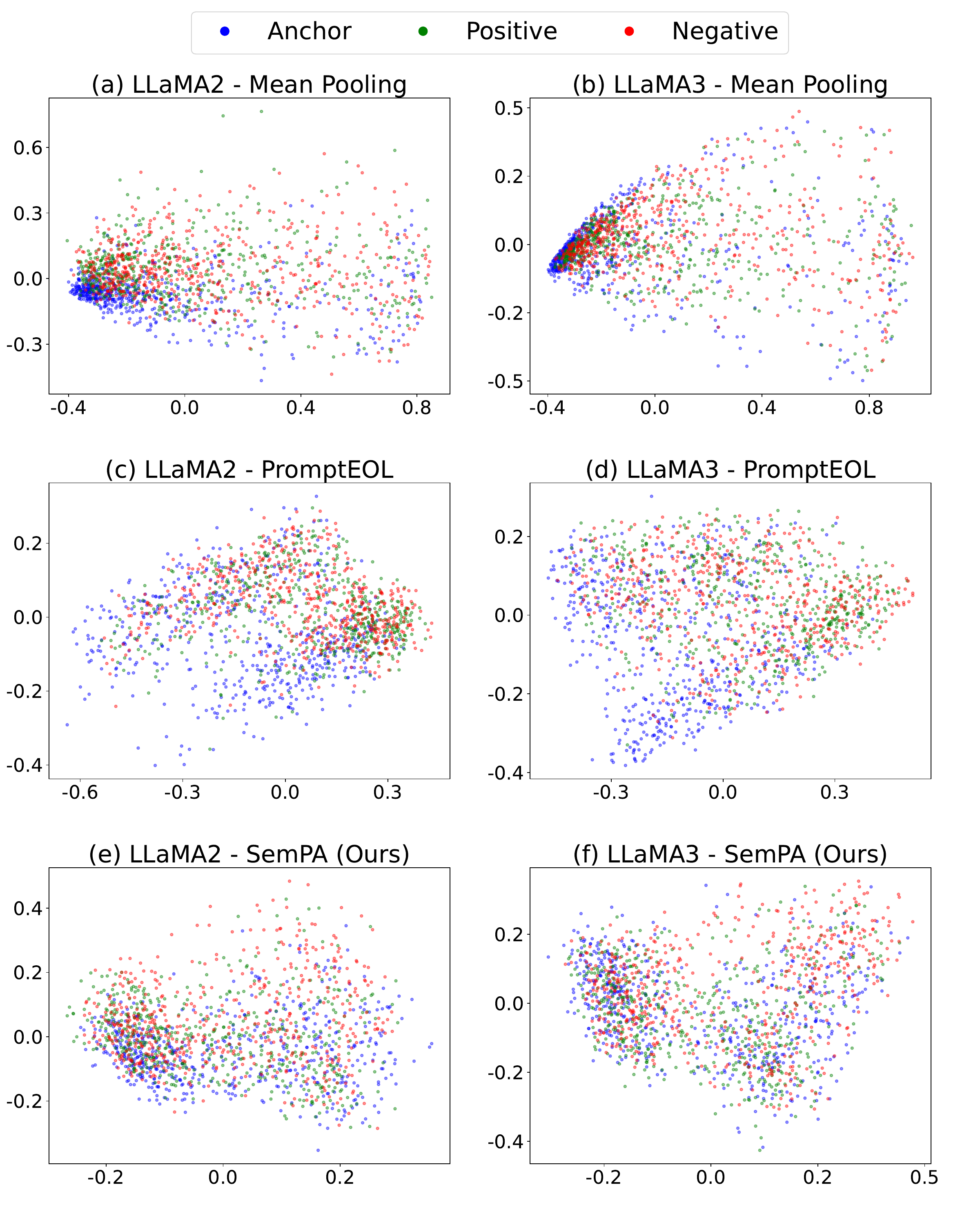}
    }
    \caption{Qualitative visualization of embedding space using LLaMA2 (left) and LLaMA3 (right). Our method results in a more isotropic embedding space.}
    \label{fig:total}
\end{figure}

\noindent\textbf{Models and Implementation Details.} 
%swift is moved to appendix
Considering the reported results of the aforementioned baselines, we use LLaMA2-7B~\cite{touvron2023llama2openfoundation} and LLaMA3-8B~\cite{grattafiori2024llama3herdmodels} as base models. We apply LoRA~\cite{hu2022lora} fine-tuning on four NVIDIA RTX 5090 GPUs to train our models. See Appendix~\ref{app:moredetails} for details.

\begin{table*}[t!]
\centering
\scalebox{0.65}{
\begin{tabular}{lcc} 
\toprule
\multirow{2.5}*{\textbf{Template $\mathcal{T}^{\rm para}$ for Training}} & \multicolumn{2}{c}{\textbf{Template $\mathcal{T}$ for Embedding Extraction}}\\
\cmidrule(lr){2-3} 
&\textbf{PromptEOL}&\textbf{Pretended CoT}\\ 
\midrule
\textit{Keep the same meaning of this sentence: ``$x^*$'', while making some changes.} & \textcolor{teal}{77.69}/\textcolor{teal}{78.10}/\textcolor{teal}{77.90}&\textcolor{magenta}{76.69}/\textcolor{teal}{78.96}/\textcolor{teal}{77.83}\\
%&\textbf{76.42}/76.45/\textbf{76.44} \\
\textit{Generate a paraphrase of this sentence: ``$x^*$'' that preserves its meaning, while making some changes.} & \textcolor{teal}{76.88}/\textcolor{teal}{78.20}/\textcolor{teal}{77.54}&\textcolor{magenta}{76.32}/\textcolor{teal}{79.00}/\textcolor{teal}{77.66}\\
%&76.37/75.10/75.74 \\
\textit{Rewrite the sentence: ``$x^*$'' while preserving its main meaning, but the wording may be simplified or rephrased.} & \textcolor{teal}{76.75}/\textcolor{teal}{75.77}/\textcolor{teal}{76.26}&\textcolor{magenta}{75.21}/\textcolor{teal}{77.90}/\textcolor{teal}{76.56}\\
%&75.66/73.61/74.64\\
\textit{Keep the main meaning of this sentence: ``$x^*$'', and rewrite it in a different way.} & \textcolor{teal}{73.62}/\textcolor{teal}{78.84}/\textcolor{teal}{76.23}&\textcolor{magenta}{71.95}/\textcolor{teal}{79.78}/\textcolor{magenta}{75.87}\\
%&69.55/\textbf{77.02}/73.29 \\
\textit{Generate a paraphrase of the sentence: ``$x^*$''.} & \textcolor{teal}{74.62}/\textcolor{teal}{76.73}/\textcolor{teal}{75.68}&\textcolor{magenta}{72.27}/\textcolor{teal}{77.84}/\textcolor{magenta}{75.06}\\
%&71.74/76.13/73.94 \\
\midrule
Baseline \textit{w/o} using $\mathcal{T}^{\rm para}$ for \textsc{SemPA} training & 70.03/71.90/70.97 & {76.86}/76.09/76.48 \\
%& 77.14/76.92/77.03\\
\bottomrule 
\end{tabular}
}
\caption{Performance (LLaMA2/LLaMA3/average) using different paraphrasing templates $\mathcal{T}^{\rm para}$ and CoT-enhanced embedding extraction template $\mathcal{T}$ of pretended CoT~\cite{10.1007/978-981-97-5669-8_5}. Apart from a decline for LLaMA2 using CoT-based template, our method consistently achieves improvements across different templates and model settings.}
\label{tab:compare_prompt}
\end{table*}

\begin{table}[t!]
\scalebox{0.71}{
\begin{tabular}{lcccc}  % l=左对齐，r=右对齐（数值列更美观）
\toprule
\textbf{Models} & \textbf{Methods} & \textbf{Uniformity}$\downarrow$ & \textbf{Isotropy Score}$\uparrow$\\
\midrule
& Mean Pooling & -0.7596 & 0.4284 \\
LLaMA2-7B & PromptEOL & -1.0066 & 0.4173 \\
& \textbf{\textsc{SemPA} (Ours)} & \textbf{-1.9139} & \textbf{0.4882} \\
\midrule
& Mean Pooling & -0.8112 & 0.4130 \\
LLaMA3-8B & PromptEOL & -1.3277 & 0.4448 \\
& \textbf{\textsc{SemPA} (Ours)} & \textbf{-2.3298} & \textbf{0.5312} \\
\bottomrule
\end{tabular}}
\caption{Quantitative results on uniformity and isotropy scores. The results are aligned with qualitative ones.}
\label{tab:align_uniform}
\end{table}

\subsection{Main Results}

The results are shown in Table~\ref{tab:sts-results}. Overall, our method achieves the highest score on 5 out of 7 datasets with LLaMA2-7B, and on 6 out of 7 datasets when leveraging the LLaMA3-8B backbone, indicating that our proposed semantic preference alignment method can indeed enhance the semantic representation capability of generative large language models.

Our method, based on LoRA fine-tuning (with only 0.3\% of the 7B model parameters being trainable), outperforms smaller-scale encoder models such as SBERT. This indicates that, in terms of performance, generative semantic representation can effectively replace traditional representation schemes. On the other hand, compared with recent LLM-based approaches such as Token Prepending, our method achieves performance improvements of 0.5$\sim$2.8 points without requiring modifications to the model architecture or computation process.

Across different backbone LLMs, we find that our method performs better with the more advanced LLaMA3-8B than with LLaMA2-7B, showing more substantial improvements over the baselines. This indicates that our approach can further exploit the intrinsic semantic representation capability as the base model's capacity increases.

\section{Analyses}
\label{section:analysis}
\noindent\textbf{Impact of Data Size.} 
Figure~\ref{fig:SP} shows the results obtained using different amounts of training data for a simple epoch. We find that both LLaMA2 and LLaMA3 achieve their highest Spearman correlation scores when trained with 40K to 80K NLI samples, and the best scores for both exceed 77. When trained with 100K or more samples, performance declines slightly and then plateaus, indicating some degree of over-alignment in paraphrase generation. Nonetheless, performance remains mostly above 73, outperforming the PromptEOL baseline. Overall, these results demonstrate that our method effectively enhances the semantic representations of generative LLMs in a data-efficient manner, without relying on large-scale, high-quality data for contrastive learning or instruction tuning.

\noindent\textbf{Visualization of Embedding Space.} 
\label{section:visualize_embedidng}
Representations of neural models often exhibits an anisotropic distribution, especially for generative LLMs, which limits their semantic expressiveness~\cite{ethayarajh-2019-contextual,li-etal-2020-sentence}. Figure \ref{fig:total} shows the PCA visualization of embeddings from randomly selected 0.2\% instances from the NLI dataset.

Results show that sentence embeddings derived via mean pooling suffer from a severe anisotropy issue, with the majority of sentence vectors clustered in a specific region of the embedding space. In comparison, PromptEOL and our \textsc{SemPA} mitigate this issue to a certain extent and generates embeddings of superior quality, where positive and negative samples are more clearly separated.

\newcommand{\tokR}[1]{\colorbox{red!35}{\smallstrut #1}}
\newcommand{\tokY}[1]{\colorbox{yellow!30}{\smallstrut #1}}
\newcommand{\tokG}[1]{\colorbox{green!40}{\smallstrut #1}}
\newcommand{\tokB}[1]{\colorbox{blue!40}{\smallstrut #1}}

We further use two metrics that assess the representation quality for comparison: 1) \textit{uniformity} introduced by~\citet{pmlr-v119-wang20k}, and 2) \textit{isotropy score} proposed by~\citet{mu2018allbutthetop}. Details of these two metrics are shown in Appendix~\ref{app:metrics}. Results in Table~\ref{tab:align_uniform} demonstrate that \textsc{SemPA} outperforms both the original model and the PromptEOL baseline across the two metrics, indicating that \textsc{SemPA} can enhance semantic expressiveness by improving the model's representation space.

\begin{table*}[t!]
\centering
\small
\setlength{\tabcolsep}{3pt}
\renewcommand{\arraystretch}{1.2} 
\begin{tabular}{@{\extracolsep{\fill}}lcccccc@{}}
\toprule
\multirow{2}*{\textbf{Models}} & \textbf{GSM8K} & \textbf{MMLU} & \textbf{HellaSwag} & \textbf{DROP} & \textbf{TruthfulQA} & \multirow{2}*{\textbf{Avg.}} \\
 & {Acc.} &{Acc.} & {Acc\_norm} & {F1} & {MC1} &  \\
\midrule
LLaMA2-7B & 14.03 & 45.91 & 76.13 & 40.23 & 24.85 & 40.23 \\
\textit{w/} Contrastive Learning & 3.79~\textcolor{magenta}{($\downarrow$10.24)} & 46.12~\textcolor{teal}{($\uparrow$0.21)} & 73.50~\textcolor{magenta}{($\downarrow$2.63)} & 15.92~\textcolor{magenta}{($\downarrow$24.31)} & 23.87~\textcolor{magenta}{($\downarrow$0.98)} & 32.64~\textcolor{magenta}{($\downarrow$7.59)}\\
\textbf{\textit{w/} Semantic DPO (Ours) }& 11.68~\textcolor{magenta}{($\downarrow$2.35)} & 47.43~\textcolor{teal}{($\uparrow$1.52)} & 79.99~\textcolor{teal}{($\uparrow$3.86)} & 31.02~\textcolor{magenta}{($\downarrow$9.21)} & 35.62~\textcolor{teal}{($\uparrow$10.77)} & 41.15~\textcolor{teal}{($\uparrow$0.92)} \\
\midrule
LLaMA3-8B & 55.80 & 64.96 & 79.22 & 58.99 & 27.29 & 57.25 \\
\textit{w/} Contrastive Learning & 9.86~\textcolor{magenta}{($\downarrow$45.94)} & 62.68~\textcolor{magenta}{($\downarrow$2.28)} & 75.44~\textcolor{magenta}{($\downarrow$3.78)} & 25.02~\textcolor{magenta}{($\downarrow$33.97)} & 24.72~\textcolor{magenta}{($\downarrow$2.57)} & 39.54~\textcolor{magenta}{($\downarrow$17.71)} \\
\textbf{\textit{w/} Semantic DPO (Ours) }& 52.01~\textcolor{magenta}{($\downarrow$3.79)} & 64.71~\textcolor{magenta}{($\downarrow$0.25)}  & 82.56~\textcolor{teal}{($\uparrow$3.34)} & 60.93~\textcolor{teal}{($\uparrow$1.94)} & 36.47~\textcolor{teal}{($\uparrow$9.18)} & 59.34~\textcolor{teal}{($\uparrow$2.09)}  \\
\bottomrule
\end{tabular}
\caption{Comparison of general capabilities of LLMs. For the results of contrastive learning, we obtain the model by extracting sentence embeddings derived from the PromptEOL template and performing contrastive learning.  }
\label{tab:generative-capacity}
\end{table*}

\begin{table}[t!]
\scalebox{0.56}{
\begin{tabular}{l} 
\toprule
\textbf{Input}: \textit{Fire from the American ships did not actually stop at this point,
and after}  \\
\textit{clarifying the position of his ships, he ordered TF 64
to resume firing at 23:51.}\\
\textbf{PromptEOL}: \tokY{The} \tokR{<0x0A>} \tokY{\_He} \tokY{Fire} \tokY{He} \tokY{\_The} \tokY{After} \tokY{after} \tokG{\_Fire} \tokR{...} \\ 
\textbf{\textsc{SemPA}}: \tokY{\_After} \tokY{res} \tokY{After} \tokG{\_res} \tokG{\_firing} \tokG{\_after} \tokY{after} \tokG{\_Fire} \tokY{Fire} \tokY{\_Res}\\ 
\midrule
\textbf{Input}: \textit{After the resumption of National College Entrance Examination, he earned}  \\
\textit{his Master of Engineering degree from the University of the Chinese Academy of}\\
\textit{Sciences in August 1981.}\\
\textbf{PromptEOL}: \tokG{\_} \tokY{\_He} \tokY{He} \tokG{\_After} \tokY{After} \tokG{\_he} \tokY{he} \tokR{<0x0A>} \tokR{<0xE5>} \tokR{<0xE6>} \\
\textbf{\textsc{SemPA}}: \tokY{After} \tokG{\_After} \tokY{\_after} \tokY{after} \tokY{Master} \tokY{August} \tokG{\_he} \tokG{\_earned} \tokY{he} \tokY{master}\\ 
\midrule
\textbf{Input}: \textit{He has been a member of the Portfolio Committee on International Relations}  \\
\textit{and Cooperation (National Assembly Committees) since 27th June 2019.}\\
\textbf{PromptEOL}: \tokG{\_He} \tokY{He} \tokR{<0x0A>} \tokY{\_he} \tokY{he} \tokR{H} \tokY{Member} \tokG{\_} \tokY{since} \tokR{I} \\
\textbf{\textsc{SemPA}}: \tokY{Member} \tokG{\_has} \tokY{\_he} \tokG{\_member} \tokG{\_He} \tokY{\_Member} \tokG{\_a} \tokY{has} \tokY{member} \tokY{He}\\ 
\midrule
\textbf{GAR Score of \textbf{PromptEOL}}: 0.185 \quad \quad \quad \quad \quad \textbf{GAR Score of \textsc{SemPA}}: 0.324\\

\bottomrule
\end{tabular}}
\caption{Top-10 aligned tokens in vocabulary according to \citet{nie-etal-2025-text}. Tokens {in green} strictly match the original input. Tokens {in yellow} match the input after stemming or lemmatization. Tokens {in red} are unrelated.}
\label{tab:aligned_tokens}
\end{table}

\noindent\textbf{Impact of Templates.}
\label{subsec:template}
We test the performance using different paraphrase generation templates $\mathcal{T}^{\rm para}$ and the embedding extraction templates $\mathcal{T}$. To guide LLMs in generating paraphrases, we design multiple instructions that include phrases like “Generate a paraphrase ...”, “Keep the meaning ...”, or “Rewrite the sentence ...”, supplemented by constraints to promote syntactic diversity. For embedding extraction, besides the PromptEOL baseline, we also use a chain-of-thought enhanced template Pretended CoT~\cite{10.1007/978-981-97-5669-8_5} as follows:

\begin{quote}
\textit{After thinking step by step, this sentence: ``$\mathcal{S}$'' means in one word:``}
\end{quote}

The averaged results on seven STS datasets using LLaMA2 and LLaMA3 are shown in Table~\ref{tab:compare_prompt}. We observe that when using PromptEOL as the sentence embedding extraction method, all five paraphrase generation templates we designed lead to significant improvements in model performance, elevating the baseline scores from 70.03/71.90 to a minimum of 74.62/76.73 and a maximum of 77.69/78.10. This demonstrates that our semantic alignment approach is generally stable and effective across different templates.

When Pretended CoT is used as the sentence embedding extraction template, we observe varying degrees of performance decline when fine-tuning the LLaMA2 model. For instance, the impact is minimal with the first two templates, while the latter three templates show a drop of 1$\sim$4 points. On the more advanced LLaMA3 model, however, our method consistently improves performance across all paraphrase generation templates. This discrepancy may stem from differences in the semantic representations elicited by CoT prompting across different base models, which could be a direction worth further exploration in future work.

\noindent\textbf{Top-10 Aligned Tokens in Vocabulary.} 
\citet{nie-etal-2025-text} finds that the tokens mapped by the embeddings from PromptEOL tend to align with key tokens from the original inputs, which helps explain the effectiveness of sentence embeddings from generative LLMs. Following their analysis, we show examples of the top-10 aligned tokens for three different texts using the LLaMA2-7B backbone in Table~\ref{tab:aligned_tokens}. These cases show that the embeddings of PromptEOL still align with some unrelated and meaningless tokens such as \texttt{<0x0A>}, which may be due to the lack of semantic refinement. In contrast, after lightweight semantic preference alignment, our method makes the aligned tokens more accurate. The GAR scores (see detailed definition in Appendix~\ref{app:gar}) proposed by \citet{nie-etal-2025-text} also show that our method indeed increases overall coverage of tokens at the dataset level.

\noindent\textbf{Impact to Generation Capability of LLMs.} 

To verify that the model fine-tuned with our proposed method can enhance its semantic representation ability while preserving its inherent generative capability, we adopt a variety of benchmarks across multiple dimensions, including mathematical reasoning (\textbf{GSM8K}; \citealp{cobbe2021trainingverifierssolvemath}), multitask language understanding (\textbf{MMLU}; \citealp{hendrycks2021measuring}), commonsense reasoning (\textbf{HellaSwag}; \citealp{zellers-etal-2019-hellaswag}), reading comprehension (\textbf{DROP}; \citealp{dua-etal-2019-drop}), and response reliability (\textbf{TruthfulQA}; \citealp{lin-etal-2022-truthfulqa}). 

The results are shown in Table \ref{tab:generative-capacity}. We find that after token-level optimization based on contrastive learning, the model's performance declines across almost all tasks (by 7.59 on LLaMA2-7B and 17.71 on LLaMA3-8B). This indicates that contrastive learning methods tend to cause the model to degenerate into a purely representational model, thereby sacrificing the original generative capabilities of the language model.

In contrast, our preference alignment approach remains a lightweight, sentence-level optimization method that does not significantly affect performance on these general tasks. Moreover, we observe that on the TruthfulQA dataset, our semantically aligned optimization improves the accuracy of the base model by 9.18$\sim$10.77 points. This reflects the inherent impact of semantic representation on certain downstream tasks.

\section{Conclusion}
We introduce \textsc{SemPA}, a sentence-level semantic preference alignment method to obtain better sentence embeddings from LLMs. We establish a theoretical connection between contrastive learning and our preference alignment method for improving sentence representations. Empirically, we evaluate \textsc{SemPA} and other approaches on a series of semantic textual similarity tasks and LLM-related benchmarks. The results show that \textsc{SemPA} enables better sentence embeddings compared with various baselines, without sacrificing the intrinsic generative capabilities of LLMs. Analysis of the embedding space and the aligned tokens of sentences shows that \textsc{SemPA} can alleviate the embedding space anisotropy and reflect the information in sentences more accurately.

\section*{Limitations}
This study preliminarily demonstrates through lightweight experiments that semantic preference optimization enhances the semantic representation of generative LLMs. Further improvements and exploration could be pursued in terms of the quantity and quality of the dataset, or automatic prompt optimization methods to enhance robustness. Additionally, our method employs pairwise DPO for preference optimization, which only considers binary comparisons between two responses. Recent advances such as LiPO~\cite{liu-etal-2025-lipo} demonstrate that listwise preference optimization can better leverage the complete ranking information among multiple candidates through sophisticated weighting mechanisms. Future work could integrate listwise ranking objectives to capture more nuanced semantic relationships.
\bibliography{custom}

\appendix
\section{More Implementation Details}
\label{app:moredetails}
% \subsection{Training Configuration} 
% \label{app:train-config}
We implement our DPO training using the {SWIFT} framework~\cite{Zhao_Huang_Hu_Wang_Mao_Zhang_Jiang_Wu_Ai_Wang_Zhou_Chen_2025}. We use four NVIDIA RTX 5090 GPUs to train LLMs with LoRA fine-tuning, setting LoRA modules to all linear layers except the language model head, with rank $r = 8$ and scaling factor $\alpha = 32$. Additionally, we set the per-step batch size to 8 and the gradient accumulation steps to 8, resulting in an effective batch size of 256. For optimization, we employ the AdamW optimizer with a learning rate schedule that includes a linear warmup phase with a ratio of 0.05, peaks at $1\times10^{-4}$, and is followed by a cosine annealing schedule. A checkpoint is saved every 20K training samples to enable subsequent investigations into the relationship between data size and performance.

\section{Uniformity and Isotropy Score}
\label{app:metrics}
\subsection{Uniformity} 

The uniformity metric serves as a quantitative measure for evaluating embedding quality. It assesses the extent to which the embedding distribution is roughly uniform on the unit hypersphere. Preserving such uniformity is essential for maintaining maximal information and effectively utilizing the feature space, attributes that have been shown to correlate strongly with performance on downstream tasks.

\noindent\textbf{Theoretical Basis.}
Given the data distribution $p_{\text {data}}$, it is formally defined as:
\begin{equation}
 \ell_{\text {uniform}} \triangleq \log \mathop{\mathbb{E}}_{\substack{x, y \sim p_{\text {data}} \\ \text{i.i.d.}}} e^{-2 \lVert f(x) - f(y) \rVert^2}.
\end{equation}

\noindent\textbf{Interpretation.}
A lower value of $\ell_{\text {uniform}}$ indicates that the embeddings are more spread out and isotropic. This uniformity is desirable because it ensures that the feature space is utilized to its full capacity. When features are uniformly distributed, they are more likely to be linearly separable and contain maximal entropy, which corresponds to preserving the most information from the data.

\subsection{Isotropy Score}
The isotropy score provides a quantitative measure of how uniformly distributed word embeddings are in the high-dimensional space. It is rigorously defined based on the partition function $Z(c)$, which aggregates the contributions of all sentence vectors in a specific direction $c$.

\noindent\textbf{Rigorous Definition.}
Given a set of sentence representations $\{v(s) : s \in V\}$, where $v(s) \in \mathbb{R}^d$, the partition function for a unit vector $c$ is defined as:
\begin{equation}
    Z(c) = \sum_{s \in V} \exp(c^\top v(s)).
\end{equation}

The isotropy score $I(\{v(s)\})$ is then defined as the ratio of the minimum to the maximum value of this partition function over all possible unit vectors:
\begin{equation}
I(\{v(s)\}) = \frac{\min_{\lVert c\rVert=1} Z(c)}{\max_{\lVert c\rVert=1} Z(c)}.
\end{equation}

Intuitively, if the embeddings are perfectly isotropic, $Z(c)$ should be a constant independent of the direction $c$, implying $I = 1$, while a score closer to 0 indicates a stronger anisotropy.

\noindent\textbf{Empirical Verification via Eigenvectors.}
Since there is no closed-form solution for $\arg\max_{\lVert c\rVert=1} Z(c)$ or $\arg\min_{\lVert c\rVert=1} Z(c)$, we employ an empirical verification strategy based on the eigenvectors of the covariance matrix. We restrict the candidate directions $c$ to the eigenvectors of $V^\top V$, where $V$ is the matrix of all sentence vectors. Let $\{u_1, \dots, u_d\}$ be the eigenvectors of $V^\top V$. The isotropy score is then estimated by computing the partition function for each eigenvector, denoted as $Z(u_j) = \sum_{s \in V} \exp(u_j^\top v(s))$, and taking the ratio of the minimum to maximum values observed among these eigenvectors:
\begin{equation}
    I_{\text {approx}} = \frac{\min_{j} Z(u_j)}{\max_{j} Z(u_j)}.
\end{equation}

\section{Comparison of GAR Score} 
\label{app:gar}
We adopt the global alignment rate (GAR) score proposed by \citet{nie-etal-2025-text} to measure, at the dataset level, how well the aligned tokens of the model’s embeddings cover the real tokens. Given an input sentence $s_i$, we first tokenize it using the tokenizer of the model and deduplicate the resulting token sequence to obtain the surface token set $T_{s_i}$. We then compute the top-$K_i$ aligned tokens of the sentence embedding and form the aligned token set $\hat{T}^{K_i}_{s_i}$. The intersection $\hat{T}^{K_i}_{s_i}\cap T_{s_i}$ indicates which tokens in the top-$K_i$ aligned tokens also appear in the surface token set. We perform the same procedure for every sentence in the dataset $D$ and take the union of the resulting ``hit token sets'' $\textstyle\bigcup_{i=1}^{|D|} (\hat{T}^{K_i}_{s_i} \cap T_{s_i})$.
The cardinality of this union represents the number of distinct surface token types that are hit by alignment across the dataset.
On the other hand, $\textstyle\bigcup_{i=1}^{|D|} T_{s_i}$
gives the set of all surface token types appearing in the dataset. Therefore, GAR is defined as the ratio of the sizes (cardinalities) of the two sets and measures the global coverage of surface tokens by the aligned tokens at the dataset level: 
\begin{equation}
\tag{11}
{\text {GAR}}
=
\frac{\left|\bigcup_{i=1}^{|D|}\left(\hat{T}^{K_i}_{s_i}\cap T_{s_i}\right)\right|}
{\left|\bigcup_{i=1}^{|D|} T_{s_i}\right|}.
\end{equation}

We randomly sample 10K of the 1M Wikipedia texts provided by~\citet{gao-etal-2021-simcse} as the dataset $D$, set $K=10$ for all sentences, and compute GAR for LLaMA2-7B before and after fine-tuning following the above procedure.
The GAR score increases from $0.185$ (pre-finetuning) to $0.324$ (post-finetuning). This result indicates that our method substantially improves the dataset-level coverage of surface token types by the model’s top-$K$ aligned tokens and thus increases the diversity of aligned surface tokens.

\section{Use of AI Assistants}
We used AI tools solely to assist with translation and language polishing (e.g., wording and grammar suggestions). The tools were not used to generate research conclusions, experimental results, or core technical content.

\end{document}